\documentclass{article}
\usepackage{ijcai22}

\usepackage{amsmath,amsfonts,bm}

\def\eqref#1{equation~\ref{#1}}

\def\1{\bm{1}}

\def\vm{{\bm{m}}}

\def\vt{{\bm{t}}}

\def\vx{{\bm{x}}}

\def\mC{{\bm{C}}}
\def\mD{{\bm{D}}}

\def\mI{{\bm{I}}}

\def\mO{{\bm{O}}}

\def\mQ{{\bm{Q}}}

\def\mW{{\bm{W}}}

\def\mZ{{\bm{Z}}}

\DeclareMathAlphabet{\mathsfit}{\encodingdefault}{\sfdefault}{m}{sl}
\SetMathAlphabet{\mathsfit}{bold}{\encodingdefault}{\sfdefault}{bx}{n}

\def\gE{{\mathcal{E}}}

\def\gG{{\mathcal{G}}}

\def\gN{{\mathcal{N}}}

\def\gV{{\mathcal{V}}}

\def\gX{{\mathcal{X}}}
\def\gY{{\mathcal{Y}}}

\def\sR{{\mathbb{R}}}

\newcommand{\R}{\mathbb{R}}

\usepackage{marvosym}
\usepackage{times}
\usepackage{soul}
\usepackage{url}
\usepackage[hidelinks]{hyperref}
\usepackage[utf8]{inputenc}
\usepackage[small]{caption}
\usepackage{graphicx}
\usepackage{amsmath}
\usepackage{amsthm}
\usepackage{booktabs}
\usepackage{algorithm}
\usepackage{algorithmic}
\usepackage{xcolor}
\newtheorem{definition}{Definition}

\title{Geometrically Equivariant Graph Neural Networks: A Survey}

\author{
Jiaqi Han$^{1,2}$
\and
Yu Rong$^{3}$
\and
Tingyang Xu$^3$
\And
Wenbing Huang$^1$\textsuperscript{\Letter}
\affiliations
$^1$Institute for AI Industry Research (AIR), Tsinghua University\\
$^2$Department of Computer Science and Technology, Tsinghua University\\
$^3$Tencent AI Lab\\
\textsuperscript{\Letter} Corresponding email: hwenbing@126.com
}

\begin{document}

\maketitle

\begin{abstract}

Many scientific problems require to process data in the form of geometric graphs. Unlike generic graph data, geometric graphs exhibit symmetries of translations, rotations, and/or reflections.  Researchers have leveraged such inductive bias and developed geometrically equivariant  Graph Neural Networks (GNNs) to better characterize the geometry and topology of geometric graphs. Despite fruitful achievements, it still lacks a survey to depict how equivariant GNNs are progressed, which in turn hinders the further development of equivariant GNNs. To this end, based on the necessary but concise mathematical preliminaries, we analyze and classify existing methods into three groups regarding how the message passing and aggregation in GNNs are represented. We also summarize the benchmarks as well as the related datasets to facilitate later researches for methodology development and experimental evaluation. The prospect for future potential directions is also provided.
\end{abstract}

\section{Introduction}

Many problems particularly in physics and chemistry require to process data in the form of \emph{geometric graphs}~\cite{bronstein2021geometric}. Distinct from generic graph data, geometric graphs assign each node not only a feature but also a geometric vector. For example, a molecule/protein can be regarded as a geometric graph, where the 3D position coordinates of atoms are the geometric vectors; or in a general multi-body physical system, the 3D states (positions, velocities or spins) are the geometric vectors of the particles. Notably, geometric graphs exhibit symmetries of translations, rotations and/or reflections. This is because the physical law controlling the dynamics of the atoms (or particles) is the same no matter how we translate or rotate  the molecular (or a general physical system) from one place to another.  When tackling this type of data, it is essential to incorporate the inductive bias of symmetry into the design of the model, which motivates the study of geometrically equivariant Graph Neural Networks (GNNs).



GNNs, originally proposed by~\cite{sperduti1997supervised}, have demonstrated their prominence in modeling graph structures under the umbrella of recent advancements of deep learning~\cite{hamilton2017inductive,huang2018adaptive,Rong2020DropEdge,sanchezgonzalez2019hamiltonian}. While abundant architectures have been developed, most previous GNNs are not geometrically equivariant\footnote{GNNs are always permutation equivariant but not inherently geometrically equivariant. This paper mainly discusses the latter  unless otherwise specified.}, making them not suitable for geometric graphs.  
To achieve geometric equivariance, plenty of works have been proposed to refine the message passing and aggregation mechanism in GNNs. These works include TFN~\cite{thomas2018tensor} equivariant on the group SE(3)---a set of 3D translation and rotation transformations, LieConv~\cite{finzi2020generalizing} on Lie Group---a set of differential transformations beyond 3D translations and rotations, and EGNN~\cite{satorras2021en} on all $n$-dimensional Euclidean transformations including translations, rotations and reflections. 


Given the fruitful achievements, however, there is still not a survey paper to depict the whole picture of how equivariant GNNs are progressed. It not only prevents the external researchers from entering this domain rapidly, but could also hinder the extraction of lessons, new ideas and visions from existing papers for the  researchers who want to push the boundary further. Hence, we establish this survey to enable a complete introduction of geometrically equivariant GNNs in a systematic way, explaining the challenges they have addressed and prospecting the potential directions for future exploration. We summarize our contributions as follows.
\begin{itemize}
    \item \textbf{Easy reference}. We provide necessary mathematical preliminaries including the definitions of equivariance, group and group representation. We try to make the math part complete but concise to avoid any unnecessary notation that could confuse readers. More importantly, all typical models are introduced by a general framework in consistent notations, such that readers are able to distinguish the difference between different methods more easily. 
    \item \textbf{Novel taxonomy}. We put forward a new taxonomy to track the development path of various equivariant GNNs. By focusing on how the message passing and aggregation are represented, we categorize equivariant GNNs into three styles: irreducible representation, regular representation, and scalarization.   
    \item \textbf{Rich resources}. Besides the methodology, we investigate the application scope of existing researches. We depict an entire list of benchmarks according to the data types they use and the tasks they target on. This can be exploited as a hand-on guidance for model development, experimental evaluation and comparison. 
    \item \textbf{Future prospects}. Provided the analyses on current works and frontier progress, we discuss the potential future directions in both theoretical and practical perspectives. In particular, we prospect the four points: theoretical completeness, scalability, hierarchy, and more real-world applications and datasets.   
\end{itemize}

\section{Backgrounds}

In this section, we introduce two key factors, namely Graph Neural Networks (GNNs) and equivariance, as the preliminaries of discussing geometrically equivariant GNNs.

\subsection{Message-passing GNNs} 
\label{sec:gnn}

GNNs have been widely adopted for handling relational data. Consider a graph $\gG = (\gV, \gE)$, where $\gV$ and $\gE$ are the set of nodes and edges, respectively. Each node is assigned a node feature, denoted as $h_i$ for node $i$. We can also optionally have an edge feature $e_{ij}$ for the edge connecting node $i$ and $j$. In~\cite{gilmer2017neural}, a seminal message-passing scheme has been refined to unify the dominated GNNs into a general architecture. It iteratively conducts message computation and neighborhood aggregation for each node (or edge). In general, we have
\begin{align}
\label{eq:message}
    m_{ij} &= \psi_m\left(h_i, h_j, e_{ij} \right), \\
\label{eq:update}
    h_i' &= \psi_h\left( \{ m_{ij} \}_{j\in\gN(i)}, h_i \right), 
\end{align}
where $\gN(i)$ is the set of neighbors around node $i$ (without self-loop by default), and $\psi_m, \psi_h$ are parametric functions. 

An intriguing property of the update depicted in Eq.~(\ref{eq:message}-\ref{eq:update}) is that it satisfies permutation equivariance, as long as $\psi_h$ is permutation equivariant. Modern GNNs are indeed designed to meet this constraint~\cite{keriven2019universal,azizian2021expressive}. Several works employ GNNs on complex systems typically with geometric inputs, \emph{e.g.}, the 3D coordinates of particles. DPI-Net~\cite{li2018learning} constructs a dynamic interaction graph based on which a physics simulator is learned for particle dynamics. HRN~\cite{mrowca2018flexible} similarly builds up an interaction graph for complex objects yet in a hierarchical manner for accurate dynamics prediction. Instead of manually specifying the graph for modeling interactions, NRI~\cite{kipf2018neural} automatically infers the latent interaction graph and achieves promising performance in modeling the dynamics of multi-body particle systems as well as human motion capture. These works have extensively demonstrated the advantages of GNNs in modeling the dynamics of real-world geometric systems, owing to their capability of reducing the combinatorial complexity of input orders due to permutation equivariance. The work by~\cite{townshend2021atomd} applies 3D GNN to predict molecular property.

Nevertheless, in this paper we focus on the geometric perspective, and the equivariance we discuss is restrained to the Euclidean space instead of the permutation group. For the above methods, it is still left to discover how the geometric symmetries in 3D space could play a role. We will explain what geometric equivariance is and why we need it in the following subsection. 

\subsection{Equivariance} 

Let $\gX$ and $\gY$ be the input and output vector spaces, respectively, both of which are endowed with a set of transformations $G$: $G\times\gX\rightarrow\gX$ and $G\times\gY\rightarrow\gY$. The function $\phi: \gX\rightarrow\gY$ is called equivariant with respect to $G$ if when we apply any transformation to the input, the output also changes via the same transformation or under a certain predictable behavior. In form, we have:
\begin{definition}[Equivariance]
The function $\phi: \gX\mapsto\gY$ is $G$-equivariant if it commutes with any transformation in $G$,
\begin{eqnarray}
\label{eq:equ}
\phi (\rho_{\gX}(g)x) = \rho_{\gY}(g)\phi(x), \forall g\in G,
\end{eqnarray}
where $\rho_{\gX}$ and $\rho_{\gY}$ are the group representations in the input and output space, respectively. Specifically, $\phi$ is called invariant if $\rho_\gY$ is the identity.
\end{definition}
\begin{definition}[Group]
A group $G$ is a set of transformations with a binary operation ``$\cdot$'' satisfying these properties: ``$\cdot$'' is closed under associative composition, there exists an identity element, and each element $G$ must have an inverse. 
\end{definition}

Given the definition of groups, we provide some examples here (more details are referred to~\cite{esteves2020theoretical}):
\begin{itemize}
    \item O($n$) is an $n$-dimensional orthogonal group that consists of rotations and reflections.
    \item SO($n$) is a special orthogonal group that only consists of rotations.
    \item E($n$) is an $n$-dimensional Euclidean group that consists of rotations, reflections, and translations.
    \item SE($n$) is a special Euclidean group that consists of rotations and translations.
    \item Lie Group is a group whose elements form a differentiable manifold. Actually, all the groups above are specific examples of Lie Group.
\end{itemize}

\paragraph{Group representation} A representation of a group is an invertible linear map $\rho(g): G \mapsto \gV$ that takes as input the group element $g\in G$ and acts on a vector space $\gV$, while at the same time it is linear: $\rho(g)\rho(h) = \rho(g\cdot h), \forall g, h\in G$. For instance, a matrix representation for O($n$) is the orthogonal  matrix $\mO\in\R^n$ subject to $\mO^{\top}\mO = \mI$. The instantiation of Eq.~(\ref{eq:equ}) on O(3) becomes $\phi(\mO x)=\mO\phi(x)$ if the input and output spaces share the same representation. For translation equivariance, we have $\phi( x-\vt)=\phi(x)-\vt$ with $t\in\R^n$.




\begin{figure}[!t]
    \centering
    \includegraphics[width=0.88\linewidth]{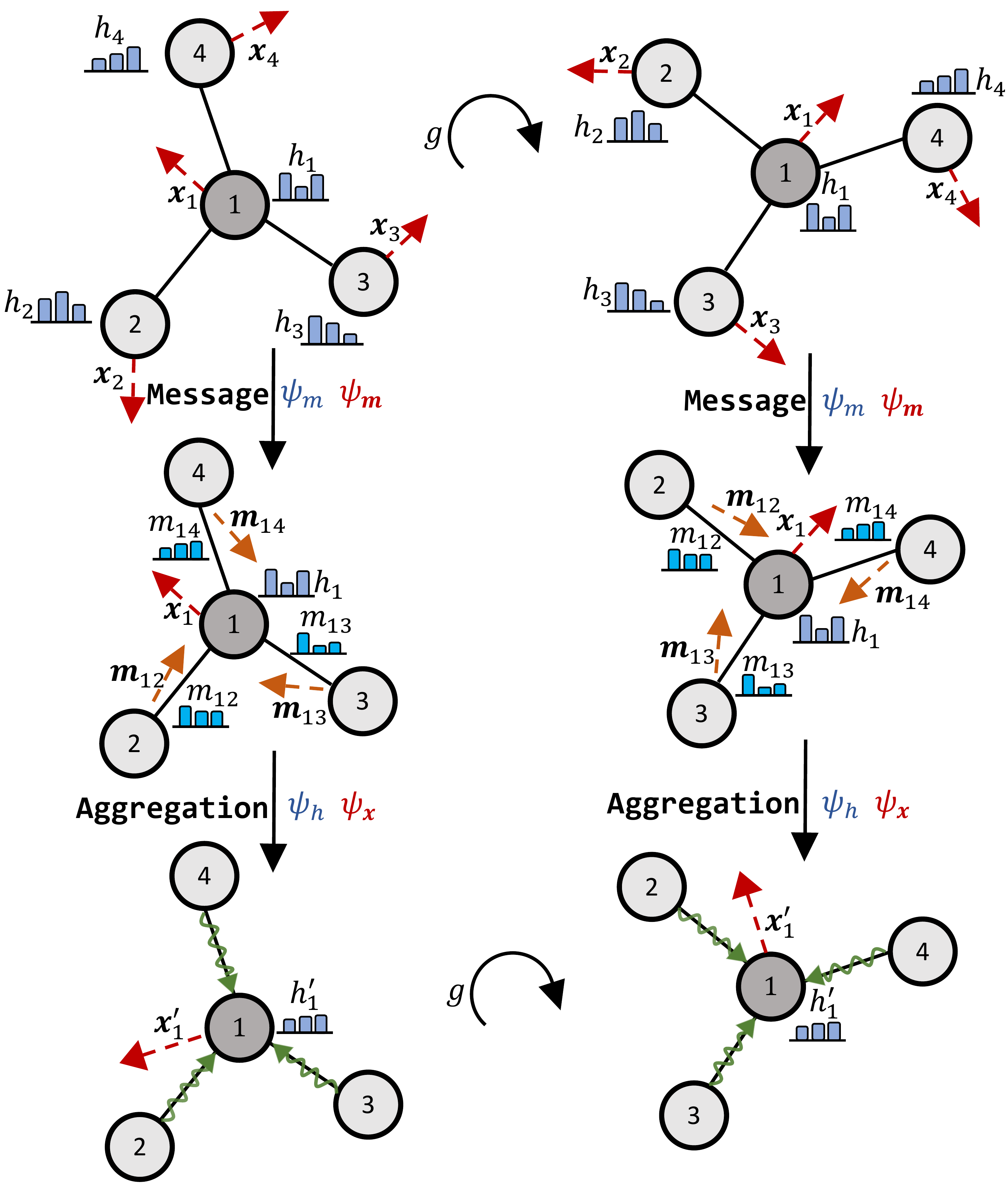}
    \caption{An illustration of the geometrically equivariant message-passing in the case of rotation. Both scalar and directional messages are produced and then aggregated, resulting in an equivariant update.}
    \label{fig:method_graph}
\end{figure}

Equivariance (\emph{a.k.a.} covariance) is initially applied to preserve symmetry in physics~\cite{anderson2019cormorant}. The endeavor of injecting Euclidean equivariance into modern deep learning frameworks originates from~\cite{cohen2016group}, where the convolution operation in CNNs has been generalized to discrete subgroups of rotations and reflections. The work by~\cite{cohen2016steerable} further introduces steerability into the convolutional kernel, generalizing the feature maps from regular representations in~\cite{cohen2016group} to irreducible and quotient representations. In what follows, we will survey how to integrate equivariance into the message passing in current GNNs.




\section{Geometrically Equivariant GNNs}
\label{sec:equiGNN}

This section first describes our formulation of geometric graphs and subsequently summarizes the geometrically equivariant GNNs that promisingly tackle this type of data.

\subsection{Geometric Graphs}

In many applications, the graphs we tackle contain not only the topological connections and node features (as illustrated in~\textsection~\ref{sec:gnn}) but also certain geometric information. Taking a molecule as an example, each atom is assigned a scalar feature $h_i \in \sR^{H}$ (\emph{e.g.}, charge, mass, etc) as well as a geometric vector $\vx_i \in \sR^{3}$ (\emph{e.g.}, position or velocity coordinates). When processing such kind of data via GNNs, we can inject the inductive bias of equivariance into the graph model. For example, when predicting the energy of a molecule, we need the output to be invariant to any rotation of the geometrical vectors; when predicting the molecular dynamics, however, we require the output to be equivariant to the input of each atom's position.  For better discrimination, we denote the geometric vector in bold and the non-geometric value via a plain symbol throughout this paper.  


In general, Eq.~(\ref{eq:message}-\ref{eq:update}) are renewed as:
\begin{align}
\label{eq:egnn-m}
    m_{ij} &= \psi_m\left(\vx_i, \vx_j, h_i, h_j, e_{ij} \right), \\
\label{eq:egnn-vm}
    \vm_{ij} &= \psi_\vm\left(\vx_i, \vx_j, h_i, h_j, e_{ij} \right),
    \\
\label{eq:egnn-h}
    h_i' &= \psi_h\left( \{ m_{ij} \}_{j\in\gN(i)}, h_i \right), 
    \\
\label{eq:egnn-x}
    \vx_i' &= \psi_\vx\left( \{ \vm_{ij}\}_{j\in\gN(i)}, \vx_i \right), 
\end{align}
where, $m_{ij}$ and $\vm_{ij}$ are separately the scalar and directional outputs by the message passing over the edge $(i,j)$, $\psi_h$ and $\psi_\vx$ are the message aggregation functions for the scalar feature and geometric vector, respectively. Besides, $\psi_m$ is $G$-invariant \emph{w.r.t.} the input $(\vx_i,\vx_j)$,  $\psi_\vm$ is $G$-equivariant \emph{w.r.t.} $(\vx_i,\vx_j)$, and $\psi_\vx$ is $G$-equivariant \emph{w.r.t.} $\{\vm_{ij}\}_{j\in\gN(i)}$ and $\vx_i$. We elaborate an illustration of the above equivariant message-passing protocol in Figure~\ref{fig:method_graph}.

Various equivariant GNNs have been proposed, and they are generally different specifications of Eq.(\ref{eq:egnn-m}-\ref{eq:egnn-x}) on different groups. We summarize them in Table~\ref{tab:models}. In terms of how the message is represented, we divide current methods into three classes: irreducible representation,  regular representation, and scalarization. In most cases, translation equivariance is trivially satisfied since the relative position $\vx_i - \vx_j$ is translation invariant, and the additive residual update in $\vx'_i$ permits translation equivariance. For this reason, our following discussions are mainly concerned with rotations and other transformations beyond translations. 

\begin{table}[htbp]
  \centering
  \small
    \setlength{\tabcolsep}{3pt}
  \caption{A summary of the equivariant GNNs. ``$\ast$'' denotes methods augmented by the attention mechanism.}
    \begin{tabular}{lccc}
    \toprule
    & Method & Group & Property \\
    \midrule
    TFN~\cite{thomas2018tensor}   & Irrep. &  SE(3)     &  Equiv.       \\
    Cormorant~\cite{anderson2019cormorant} & Irrep. & SO(3) & Equiv. \\
    SE(3)-Tr.~\cite{fuchs2020se}$^\ast$    & Irrep. &  SE(3)     &  Equiv.       \\
    NequIP~\cite{batzner2021e3equivariant} & Irrep. & E(3) & Equiv. \\
    SEGNN~\cite{brandstetter2022geometric}    & Irrep. &   E(3)    &   Equiv.      \\
    \midrule
    LieConv~\cite{finzi2020generalizing}    & Regul. &  Lie      &  Invar.      \\
    LieTr.~\cite{hutchinson2021lietransformer}$^\ast$    & Regul. &  Lie      &  Invar.      \\
    \midrule
    SchNet~\cite{schutt2018schnet} & Scala. & E(3) & Invar.  \\
    DimeNet~\cite{Klicpera2020Directional} & Scala. & E(3)  & Invar.  \\
    SphereNet~\cite{liu2022spherical} & Scala. & SE(3) & Invar. \\
    Radial Field~\cite{kohler2020equivariant} & Scala. & E($n$) & Equiv. \\
    GVP-GNN~\cite{jing2021learning} & Scala. & E(3) & Equiv. \\
    EGNN~\cite{satorras2021en} & Scala. & E($n$)  & Equiv.  \\
    GMN~\cite{huang2022constrained}  & Scala. & E($n$) & Equiv.  \\
    PaiNN~\cite{pmlr-v139-schutt21a}  & Scala. & SO(3) & Equiv.  \\
    ET~\cite{tholke2022equivariant}$^\ast$ & Scala. & O(3)  & Equiv. \\
    GemNet~\cite{klicpera2021gemnet} & Scala. & SE(3) & Equiv.  \\
    \bottomrule
    \end{tabular}%
  \label{tab:models}%
\end{table}%

\subsection{Irreducible Representation}

According to representation theory~\cite{esteves2020theoretical}, the linear representations of a compact group can be expressed by the direct sum of \emph{irreducible representations} (or \emph{irreps} for short) up to a similarity transformation. Specifically for the group SO(3), the irreps are $(2l+1)\times(2l+1)$ Wigner-D matrices $\mD^l$ with non-negative integer $l=0,1,\cdots$. For every SO(3) representation, we have
\begin{align}
    \label{eq:irreps}
    \rho(g) = \mQ^{\top} \left(\bigoplus\limits_{l} \mD^l(g)\right) \mQ,
\end{align}
where $\mD^l$ is the Wigner-D matrix, $\mQ$ is an orthogonal matrix accounting for the change of basis, and $\bigoplus$ is the direct sum or concatenation of matrices along the diagonal. By this means, the vector space is partitioned into $l$ subspaces, each transformed by $\mD^l$, and the vector lies in the $l$-th subspace is dubbed type-$l$ vector. For instance, in our case the scalar $h_i$ is a type-$0$ vector with $H$ channels, and $\vx_i$ is a type-$1$ vector. These vectors interact via the tensor product ``$\otimes$'', and the tensor product of Wigner-D matrices yields the Clebsch-Gordan (CG) coefficients $\mC^{lk}\in\R^{(2l+1)(2k+1)\times(2l+1)(2k+1)}$ via the CG decomposition:
\begin{align}
    \label{CGdecomposition}
    \mD^k(g)\otimes \mD^l(g) = (\mC^{lk})^\top\left(\bigoplus\limits_{J=|k-l|}^{k+l}\mD^J(g)  \right) \mC^{lk}.
\end{align}
The last recipe for building an equivariant message passing layer is the spherical harmonics $Y_{Jm}$ that serves as an equivariant basis for SO(3). With the above building blocks, ~\cite{thomas2018tensor} proposes the TFN layer satisfying SE(3)-equivariance,

\begin{align}
\label{eq:tfn}
    \vm_{ij}^l &= \sum_{k\geq0}\mW^{lk}(\vx_i-\vx_j)\vx^k_j, \\
    \vx'^{l}_i &= w_{ll}\vx^l_i + \sum_{j\in\gN(i)}\vm^l_{ij},
\end{align}
where $\vx^l_i\in\R^{2l+1}$ denotes the geometric vector of degree $l$ for node $i$, $\vx_i\in\R^3$ is the node coordinate, $w_{ll}$ is the self-interaction weight, and the filter $\mW^{lk}(\vx)\in\R^{(2l+1)\times(2k+1)}$ is rotation-steerable, implying $\mW^{lk}(\mD^1(r)\vx)=\mD^l(r)\mW^{lk}(\vx)(\mD^{k}(r))^{-1}$ for an arbitrary rotation $r\in\text{SO(3)}$. To be specific, $\mW^{lk}(\vx)=\sum_{J=|l-k|}^{l+k}\varphi^{lk}_J\left(\|\vx\|\right)\sum\nolimits_{m=-J}^JY_{Jm}(\vx/\|\vx\|)\mC^{lk}_{Jm}$, a combination of learnable  radius functions $\varphi^{lk}_J\in\R$,  spherical harmonics $Y_{Jm}\in\R$, and CG coefficients $\mC^{lk}_{Jm}\in\R^{(2l+1)\times(2k+1)}$. More details can also referred to~\cite{weiler20183d}.

TFN regards $\vx^l_i$ as a signal function of $\vx_i$, and the computations above only update $\vx^l_i$ while keeping $\vx_i$ fixed over all layers. It is easy to check the equivariance of $\vx^l_i$ with the help of the steerability of  $\mW^{lk}(\vx)$. TFN respectively realizes the general form in Eq.~(\ref{eq:egnn-m})\&(\ref{eq:egnn-h}) by setting $l=0$, and Eq.~(\ref{eq:egnn-vm})\&(\ref{eq:egnn-x}) by setting $l=1$ if $\vx_i$ is considered as a function of itself. 

Several extensions have been made since TFN. \cite{fuchs2020se} further incorporate the attention mechanism to Eq.~\ref{eq:tfn} by multiplying $m^l_{ij}$ with an SE(3)-invariant attentive term. \cite{dym2021on} theoretically reveals that both TFN~\cite{thomas2018tensor} and SE(3)-Transformer~\cite{fuchs2020se} are universal approximators for SE(3)-equivariant functions. Cormorant~\cite{anderson2019cormorant} similarly leverages the irreps but with Clebsch-Gordan non-linearities. NequIP~\cite{batzner2021e3equivariant} further lifts to the E(3) symmetry for the prediction of interatomic potentials. Recent work by~\cite{brandstetter2022geometric} generalizes to steerable vectors and builds a competitive E(3)-equivariant GNN with steerability. However, the computational overhead of these methods is generally expensive, restricting the usage of vectors from higher degrees. Moreover, the irreps are only widely known for groups like SO(3), while not implementation-friendly to many other groups, \emph{e.g.}, E($n$).

\subsection{Regular Representation}

Another line of work directly seeks the solution to obtaining equivariance from group convolution~\cite{cohen2016group} using regular representation, which defines convolution filters as functions on groups. Nevertheless, the integral in group convolution becomes intractable when dealing with continuous and smooth groups, and one feasible tool to leverage in this case is Lie algebra. To this end, ~\cite{finzi2020generalizing} proposes LieConv that figures out group convolution via lifting (mapping the input in $\gX$ to a group element $g\in G$) and discretization of the convolution integral via the
PointConv trick. Particularly, with our consistent notations, LieConv is formulated as follows.
\begin{align}
\label{eq:lie-m}
    m_{ij} &= \varphi\left(\log(u_i^{-1}u_j)\right)h_j, \\
\label{eq:lie-h}
    h_i' &= \frac{1}{|\gN(i)|+1}\left(h_i + \sum_{j\in\gN(i)} m_{ij}\right),  
\end{align}
where $u_i\in G$ is a lift of $\vx_i$, the logarithm $\log$ maps each group member onto the Lie Algebra $\mathfrak{g}$ that is a vector space, and $\varphi$ is a parametric MLP. Besides, Eq.~(\ref{eq:lie-h}) conducts normalization by the division of the number of all nodes, \emph{i.e.} $\gN(i)+1$.
It is clear that LieConv only specifies the update of node features $h_i$ while keeping the geometric vectors $\vx_i$ unchanged. That means LieConv is invariant\footnote{LieConv is originally claimed as equivariant in~\cite{finzi2020generalizing}, which is explained in the sense that $h'_i$ is a signal function of $\vx'_i$ (not a function of $\vx_i$ in our case).}. 

Following a similar idea, LieTransformer~\cite{hutchinson2021lietransformer} employs the self-attention strategy to dynamically re-weight the convolutional kernel for an increase in model capacity and performance. The regular-representation-based methods enjoy higher flexibility since the equivariance can be obtained on arbitrary Lie groups as well as their discrete subgroups. On the other hand, due to discretization and sampling, they also encounter a trade-off between computational complexity and performance. Another weakness is that it is non-trivial to extend Eq.(\ref{eq:lie-m}-\ref{eq:lie-h}) for the propagation of geometric vectors, unless we introduce external Hamiltonian dynamics to renew geometric vectors akin to~\cite{finzi2020generalizing}. 

\subsection{Scalarization}

Aside from group representation theory, a number of works adopt a generic way of modeling equivariance through scalarization. Typically, the geometric vectors are firstly transformed into invariant scalars, followed by several MLPs to control the magnitude, and finally added up in the original directions to obtain equivariance. This idea has originally been implied in SchNet~\cite{schutt2018schnet} and DimeNet~\cite{Klicpera2020Directional}, but only in the invariant flavor. SphereNet~\cite{liu2022spherical} further involves angular and torsional information in the scalarized message-passing, contributing to an invariant network that can distinguish chirality. Radial Field~\cite{kohler2020equivariant} implements the equivariant version, but it only operates on geometric vectors without the consideration of node features. EGNN~\cite{satorras2021en} further refines the idea via a flexible paradigm:
\begin{align}
\label{eq:egnnmessage}
    m_{ij} &= \varphi_m\left(h_i, h_j, \|\vx_i-\vx_j\|^2,e_{ij} \right), \\
    \label{eq:egnnequ}
    \vx'_i &= \vx_i + \sum_{j\neq i}\left(\vx_i-\vx_j \right)\varphi_x(m_{ij}), \\
    \label{eq:egnninv}
    h'_i &= \varphi_h (h_i, \sum_{j\in\gN(i)}m_{ij} ),
\end{align}
where, $ \|\vx_i-\vx_j\|^2$ is the scalarization of the geometric vectors $\vx_i$ and $\vx_j$; $\varphi_m$, $\varphi_x$, and $\varphi_h$ are arbitrary MLPs. By setting $\vm_{ij}=\left(\vx_i-\vx_j \right)\varphi_x(m_{ij})$, EGNN implement Eq.~(\ref{eq:egnn-m}-\ref{eq:egnn-x}) by jointly propagating the node features $h_i$ and geometric vectors $\vx_i$ in an equivariant yet straightforward way. The essence lies in constructing the invariant message $m_{ij}$, and then transforming back to the equivariant output along the radial directions, analogous to the way we compute the resultant Coulomb (or gravitational) force exerted on pairs of charged particles. Notice that Eq.~(\ref{eq:egnnequ}), different from Eq.~(\ref{eq:egnninv}), aggregates messages from all nodes in addition to the neighbors around $i$, which reflects the law that the dynamics of each node is influenced by all others.

Beyond EGNN, GMN~\cite{huang2022constrained} extends the aggregation to deal with multiple geometric vectors (\emph{e.g.} positions and velocities plus forces), denoted by $\mZ\in\sR^{3\times m}$, showing that $\mZ' = \mZ\varphi(\mZ^{\top}\mZ)$ is indeed a universal form in this case. By this means, the interaction in Eq.~(\ref{eq:egnnmessage}-\ref{eq:egnnequ}) could span across the space with $\mZ$ as the basis instead of the radial direction alone. This is particularly desirable in constrained systems, where non-radial message vectors (\emph{e.g.}, angular momentum, torque) might be produced by the interaction between and within objects. Despite the simplicity, the theoretical efficacy of these methods is endorsed by~\cite{villar2021scalars}, stating that the scalarization (inner product) techniques are universal to achieve equivariance. GemNet~\cite{klicpera2021gemnet} leverages this universality to inject rich geometric information, for example, the dihedral angle, in the message passing, putting a step forward from the framework of DimeNet~\cite{Klicpera2020Directional}.

One interpretation of Eq.~(\ref{eq:egnnequ}) is that the multiplication of invariant scalar and equivariant vector still produces an equivariant vector. In this principle, another line of work conducts equivariant message passing in various forms. PaiNN~\cite{pmlr-v139-schutt21a} and the attentive Equivariant Transformer~\cite{tholke2022equivariant} augment the invariant SchNet into equivariant flavor by projecting the inter-atomic distances via radial basis functions and iteratively updating the vectors along with the scalar features. GVP-GNN~\cite{jing2021learning} leverages a similar idea but comes with a stronger theoretical guarantee of universality.



\begin{table*}[!t]
  \centering
  \small
  \caption{Datasets for the evaluation of equivariant GNNs. }
  \vskip-0.1in
    \begin{tabular}{lccc}
    \toprule
    Dataset & Application    & Task  & Property \\
    \midrule
    N-body~\cite{kipf2018neural}     &   Physical Simulation         &  Position \& Velocity  Prediction  &  Equivariant \\
    Constrained N-body~\cite{huang2022constrained}    &  Physical Simulation           &  Position Prediction  & Equivariant  \\
    Motion Capture~\cite{cmu2003motion}     &  Physical Simulation       & Position Prediction       & Equivariant \\
    \midrule
    QM9~\cite{ramakrishnan2014quantum}     &  Small Molecule         & Chemical Property Prediction     & Invariant   \\
    MD17~\cite{chmiela2017machine}     &  Small Molecule            & Energy \& Force Prediction  & Invariant \\
    ISO17~\cite{schutt2018schnet}     &  Small Molecule        & Energy \& Force Prediction  & Invariant \\
    OC20~\cite{zitnick2020introduction}     &  Molecule (Catalyst)       & Relaxed Energy, \emph{etc.} Prediction        & Invariant  \\
      GEOM~\cite{axelrod2022geom}  &  Small Molecule   & Generation   & Equivariant \\
    Atom3D~\cite{townshend2021atomd}     & Molecule \& RNA \& Protein        &  Binding Affinity, \emph{etc.} Prediction      & Invariant \\
    MDAnalysis~\cite{oliver_beckstein-proc-scipy-2016}     &  Protein    &  Position Prediction     & Equivariant  \\
    \midrule 
    ModelNet40~\cite{Wu_2015_CVPR} & Point Cloud   & Classification  & Invariant \\
    ScanObjectNN~\cite{Uy_2019_ICCV} & Point Cloud   & Classification & Invariant \\
    \bottomrule
    \end{tabular}%
  \label{tab:dataset}%
  \vskip-0.1in
\end{table*}%

\section{Application}

Equivariant GNNs have wide applications on various types of real-world geometric data, ranging from physical systems to chemical substances. In this section, we introduce the application in scenarios involving physical systems, molecular data, and point clouds, respectively. An overview of the resources of datasets is displayed in Table~\ref{tab:dataset}.

\subsection{Physical Dynamics Simulation}
Modeling the dynamics of complex physical systems has long been a challenging topic, and neural networks have been applied to infer the interaction and dynamics. Within the physical system are objects like charged particles that interact through forces abiding by physical laws.~\cite{kipf2018neural} introduces the N-body simulation, in which multiple charged particles are driven by Coulomb force. The task here is to predict the dynamics of the particles given the initial conditions, including positions, velocities, and charges. Such a task is E(3)-equivariant, since the dynamics of particles translate, rotate, and reflect together with the entire system. Both~\cite{fuchs2020se} and~\cite{satorras2021en} showcase the strong performance of equivariant GNNs on this task.~\cite{huang2022constrained} equips the system with connected rigid bodies like sticks and hinges, contributing to a more challenging scenario dubbed Constrained N-body.~\cite{brandstetter2022geometric} creates a counterpart of the system by instead using gravitational force and increasing the number of particles significantly.~\cite{kipf2018neural} and~\cite{huang2022constrained} also adopt human motion capture data~\cite{cmu2003motion} which involves human motion trajectories towards different directions.

\subsection{Molecules}
Another essential application scenario lies in molecular data, where atoms in molecules are affected by complicated chemical interactions. For molecular data, the scalar node feature $h_i$ is typically the atom number, and the connectivity between nodes is either provided by chemical bonds or taken by a cut-off based on a distance threshold. We review Equivariant GNNs on molecules, including prediction and generation. 

\paragraph{Prediction} QM9~\cite{ramakrishnan2014quantum} is a widely adopted dataset comprising 12 quantum properties as invariant tasks. MD17~\cite{chmiela2017machine} includes molecular dynamics trajectory for 8 small molecules alongside the energy and interaction force as labels. Similar to MD17, ISO17~\cite{schutt2018schnet} contains short MD trajectories of 129 isomers. The Open Catalyst 2020 (OC20) dataset considers the binding process of catalyst and absorbate, and the tasks here include predicting the relaxed energy or structure given the initial structures. A large number of works discussed in \textsection~\ref{sec:equiGNN} have demonstrated the efficacy on these datasets by predicting the objectives accurately while requiring less computational cost than the traditional simulation method via Density Functional Theory (DFT). 

Moving forward, AlphaFold2~\cite{AlphaFold2021} achieves striking performance in predicting the folding structure of proteins, which initiates endeavors towards large chemical compounds with the help of geometric equivariance. Several works extend the basic frameworks described in \textsection~\ref{sec:equiGNN} on large-scale molecular systems like proteins.~\cite{eismann2020hierarchical} employs TFN~\cite{thomas2018tensor} on protein data in a hierarchical manner by sampling the key substructures.~\cite{ganea2022independent} targets the problem of rigid body protein-protein docking by extending~\cite{satorras2021en} to a matching network that copes with two geometric graphs simultaneously. ARES~\cite{raphael2021geometric} applies a rotationally equivariant network on RNA data, and interestingly, the model exhibits strong generalization capability even trained with only 18 RNA structures. On the data side, MDAnalysis~\cite{oliver_beckstein-proc-scipy-2016} serves as a solid dataset for molecular dynamics on proteins. Atom3D~\cite{townshend2021atomd} is a comprehensive dataset containing 8 prediction tasks on molecules with geometric information, ranging from small molecules to RNA and proteins.

\paragraph{Generation}
\cite{axelrod2022geom} introduce a large-scale unlabeled data split into two subsets GEOM-QM9 and GEOM-Drugs. These datasets contain diverse samples of conformations for a large number of small molecules and have served as the unsupervised training set in the task of molecular conformation generation. ConfGF~\cite{shi2021learning} and DGSM~\cite{luo2021predicting} parameterize the score function in the score-based generative model~\cite{song2019generative} with a roto-translation equivariant GNN, leading to a generative model that learns the conditional distribution of the conformations. GeoDiff~\cite{xu2022geodiff} further promotes the generative model to denoising diffusion model~\cite{ho2020denoising}, and similarly adapt the diffusion kernel by a GNN with equivariance guarantee. Besides, Equivariant Flow~\cite{kohler2020equivariant} verifies the feasibility of constructing normalizing flow with an equivariant kernel in the change of density. However, only the coordinates are taken into account without the node feature.~\cite{satorras2021enf} makes use of EGNN (Eq.~\ref{eq:egnnmessage}-\ref{eq:egnninv}) as the kernel and jointly models the evolution of both vector and scalar inputs in the dynamics of continuous normalizing flow~\cite{chen2018continuous}. These methods allow for unconditional generation, where the conformations are generated from scratch without providing the molecular graph.


\subsection{Point Clouds}

Point cloud is a representation format of objects that describe the shape by a set of points assigned with coordinates. ModelNet40~\cite{Wu_2015_CVPR} and ScanObjectNN~\cite{Uy_2019_ICCV} are two well-received datasets for object classification based on the point cloud. Since the connectivity of points is absent in point cloud data, the neighborhood used by equivariant GNNs is commonly defined by $\gN(i) = \{ j \big| \|\vx_i-\vx_j\|_2 < d  \}$ where $d$ is the maximum distance. Both TFN and SE(3)-Transformer have exhibited competitive performance on point cloud data. Recently,~\cite{Chen_2021_CVPR} proposes an SE(3)-equivariant convolution scheme for point cloud by iteratively conducting group convolution and point convolution. These approaches demonstrate an advantage of equivariant GNNs compared to traditional 3D CNNs, since they do not require the voxelization of input Euclidean space while still maintaining the desirable equivariance.




\section{Future Research Directions}

\paragraph{Theoretical completeness} 
Unlike the well established theoretical frameworks for depicting the expressivity and generalization of GNNs, little has been revealed about equivariant GNNs. Although several works~\cite{dym2021on,villar2021scalars,jing2021learning} have analyzed the universality of some methods, their discussions are merely based on the message-passing function, whereas the property of the entire graph model remains unknown. It will be interesting to see how the existing theoretical framework in building powerful equivariant GNNs converge to this respect.

\paragraph{Scalability} 
As discussed in~\textsection~\ref{sec:equiGNN}, the methods leveraging group representation theory consume high computational cost, limiting their scalability towards large and complex systems like proteins. The issue becomes more significant when they are equipped with attention. As a consequence, it is still in demand of speeding up the computation in the equivariant message passing. Possible solutions to improving scalability involve approximation and efficient sampling. 

\paragraph{Hierarchical structure}
Many real-world systems exhibit hierarchical structures. For example, organic molecules are comprised of multiple functional groups, and proteins consist of amino acids. By making use of these hierarchical structures, it is possible for equivariant GNNs to model the systems in multiple granularities. Compared with the flat message passing scheme adopted by existing works, we expect the hierarchical mechanism to enhance the efficiency as well as generalization of the model.

\paragraph{More real-world applications and datasets}
Many equivariant GNNs have only been evaluated on systems with limited scale and complexity, \emph{e.g.}, N-body system and the small molecular datasets. More challenging tasks involve strengthening the systems with more objects, more diverse constraints, and more complicated interactions. Works on proteins are promising attempts, but they are mostly limited in diversity due to difficulty in the collection, and a comprehensive evaluation of existing methods has not been established. 

\section{Conclusion}

In this paper, we conduct a survey of geometrically equivariant GNNs. We show that existing works lie in our refined geometric message-passing paradigm as specifications via irreducible representation, regular representation, or scalarization. We discuss their broad application prospect on various tasks, including simulating physical dynamics, modeling molecules or proteins, and handling point clouds. With the promising future directions, we expect to see equivariant GNNs as powerful tools for coping with the tasks in scientific domains.

\section*{Acknowledgments}
This work was jointly supported by the following projects: the Scientific Innovation 2030 Major Project for New Generation of AI under Grant NO. 2020AAA0107300, Ministry of Science and Technology of the People's Republic of China; the National Natural Science Foundation of China (No.62006137)

\bibliographystyle{named}
\bibliography{ijcai22}

\end{document}